\documentclass[11pt,a4paper]{article}
\usepackage[hyperref]{emnlp-ijcnlp-2019}
\usepackage{times}
\usepackage{latexsym}

\usepackage{url}

\usepackage{graphicx} 
\usepackage{booktabs} 
\usepackage{array}
\usepackage{multirow}
\usepackage{amsmath} 
\usepackage{subfig} 

\newcolumntype{L}[1]{>{\raggedright\let\newline\\\arraybackslash\hspace{0pt}}m{#1}}
\newcolumntype{C}[1]{>{\centering\let\newline\\\arraybackslash\hspace{0pt}}m{#1}}
\newcolumntype{R}[1]{>{\raggedleft\let\newline\\\arraybackslash\hspace{0pt}}m{#1}}

\makeatletter
\DeclareRobustCommand*\textsubscript[1]{%
	\@textsubscript{\selectfont#1}}
\def\@textsubscript#1{%
	{\m@th\ensuremath{_{\mbox{\fontsize\sf@size\z@#1}}}}}
\makeatother

\newcommand{\squishlist}{
 \begin{list}{$\bullet$}
  { \setlength{\itemsep}{0pt}
     \setlength{\parsep}{3pt}
     \setlength{\topsep}{3pt}
     \setlength{\partopsep}{0pt}
     \setlength{\leftmargin}{1.5em}
     \setlength{\labelwidth}{1em}
     \setlength{\labelsep}{0.5em} } }

\newcommand{\squishlistnobullet}{
 \begin{list}{}
  { \setlength{\itemsep}{0pt}
     \setlength{\parsep}{3pt}
     \setlength{\topsep}{3pt}
     \setlength{\partopsep}{0pt}
     \setlength{\leftmargin}{0em}
     \setlength{\labelwidth}{1em}
     \setlength{\labelsep}{0.5em} } }

\newcommand{\squishend}{
  \end{list}  }

\aclfinalcopy 

\title{STANCY: Stance Classification Based on Consistency Cues}

\author{Kashyap Popat$^1$, Subhabrata Mukherjee$^2$, Andrew Yates$^1$, Gerhard Weikum$^1$ \\
	$^{1}$Max Planck Institute for Informatics, Saarbr{\"u}cken, Germany \\
	$^{2}$Microsoft Research, Redmond, WA, USA \\
	$^{1}${\tt \{kpopat,ayates,weikum\}@mpi-inf.mpg.de}\\ $^{2}${\tt subhabrata.mukherjee@microsoft.com}
}

\date{}

\begin{document}
\maketitle
\begin{abstract}
Controversial claims are abundant in online media and discussion forums. A better understanding of such claims requires analyzing them from different perspectives. Stance classification is a necessary step for inferring these perspectives in terms of supporting or opposing the claim. 
In this work, we present a neural network model for stance classification leveraging BERT representations and augmenting them with a novel consistency constraint. Experiments on the \textit{Perspectrum} dataset, consisting of claims and users' perspectives from various debate websites, demonstrate the effectiveness of our approach over state-of-the-art baselines.
\end{abstract}

\section{Introduction}

There is an abundance of contentious claims on the Web including controversial statements from politicians, biased news reports, rumors, etc. People express their perspectives about these controversial claims through various channels like editorials, blog posts, social media, and discussion forums. To achieve a deeper understanding of these claims, we need to understand users' perspectives and stance towards the claims. Recent research \cite{FNC, baly-etal-2018-integrating, CKYCR19} has shown stance classification to be a critical step for information credibility and automated fact-checking.

\begin{figure*}[t]
	\centering
	\subfloat[BERT\textsubscript{BASE}: Fine-tuning BERT for stance classification.]{%
		\includegraphics[width=0.48\linewidth]{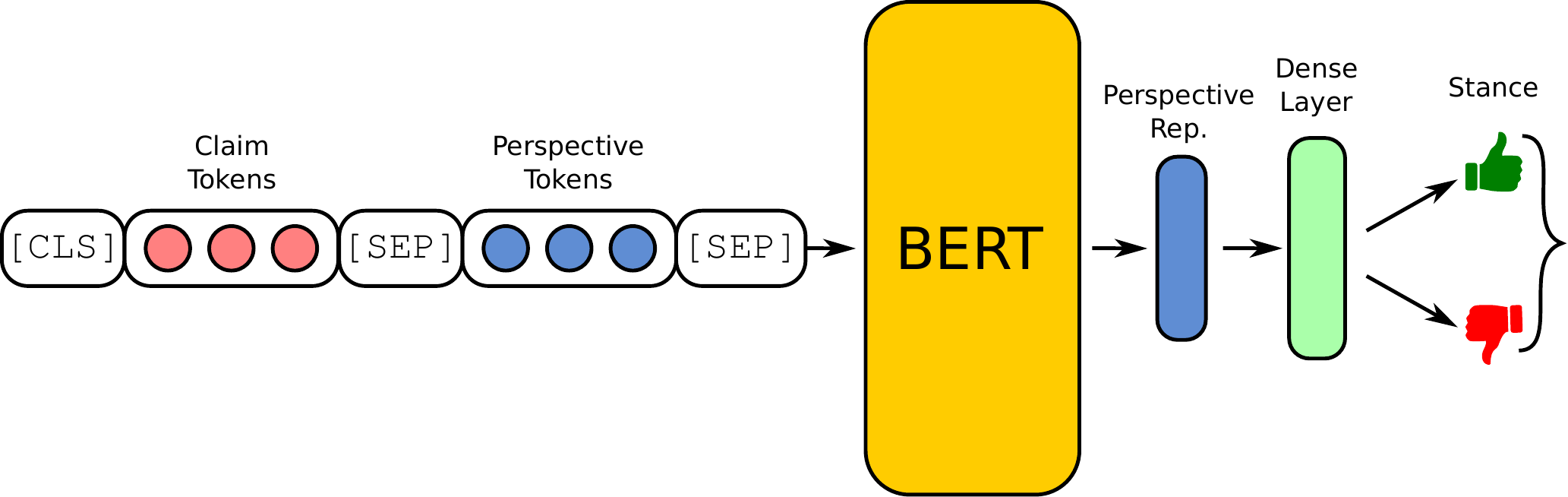}
		\label{fig:1a}
	}
	\hfill
	\subfloat[BERT\textsubscript{CONS}: Enhancing BERT using  the joint loss \hspace{0.5cm}($loss_{ce}$ for stance classification and $loss_{cos}$ for consistency).]{%
		\includegraphics[width=0.48\linewidth]{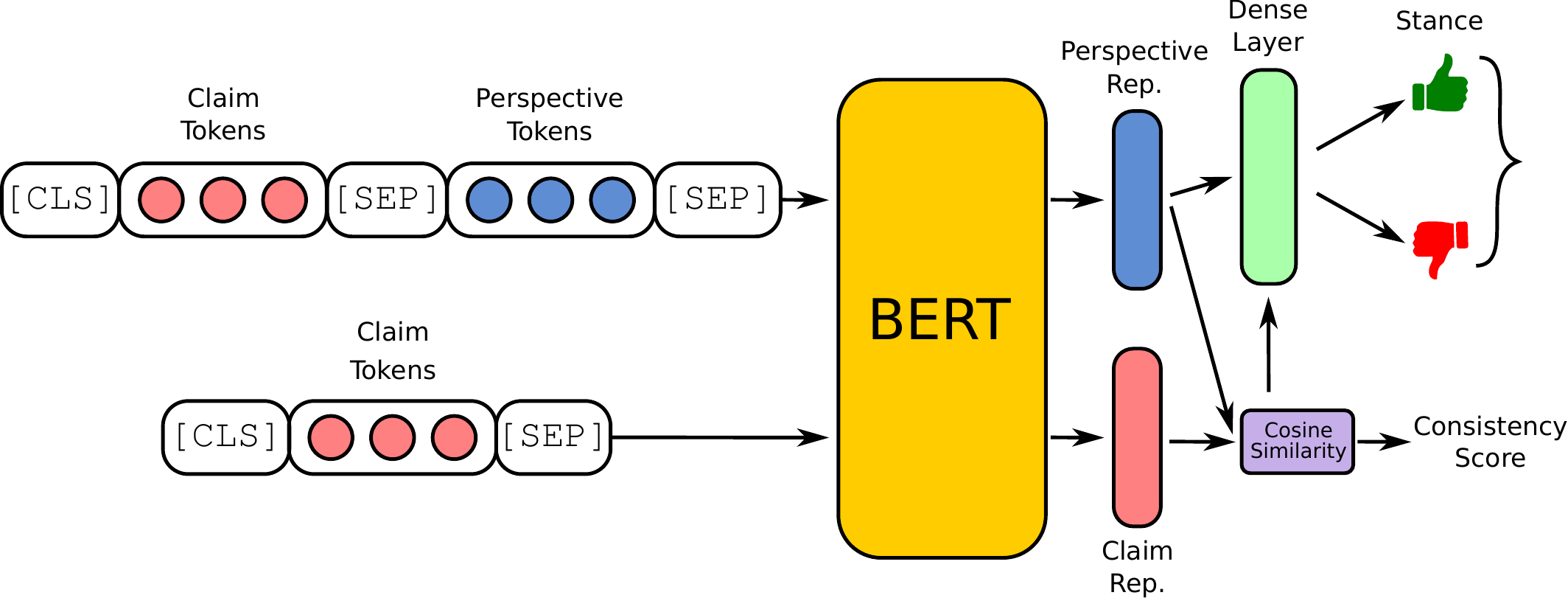}
		\label{fig:1b}
	}
	\vspace{-0.1cm}
	\caption{BERT-based methods for determining the stance of the perspective with respect to the claim. 
	}
	\label{fig:model}
	\vspace{-0.1cm}
\end{figure*}

\textbf{Prior Work and Limitations:}
Prior approaches for stance classification proposed in \citet{somasundaran-wiebe-2010-recognizing,anand-etal-2011-cats,walker-etal-2012-stance,hasan-ng-2013-stance, Hasan2014WhyAY,sridhar-etal-2015-joint,sun-etal-2018-stance}
rely on various linguistic features, e.g., n-grams, dependency parse tree, opinion lexicons, and sentiment to determine the stance of perspectives regarding controversial topics.
\citet{ferreira-vlachos-2016-emergent} further incorporate natural language claims and propose a logistic regression model using the lexical and semantic features of claims and perspectives. 
SemEval tasks \cite{mohammad-etal-2016-semeval,KochkinaLA17} and other approaches \cite{chen-ku-2016-utcnn, lukasik-etal-2016-hawkes,sobhani-etal-2017-dataset} have focused on determining stance only in Tweets. 

\citet{bar-haim-etal-2017-stance} propose classifiers based on hand-crafted lexicons to identify important phrases in perspectives and their consistency with the claim to predict the stance
However, their model critically relies on manual lexicons and assumes that the important phrases in claims are already identified.

Neural-network-based approaches for stance classification learn the claim and perspective representations separately and later combine them with conditional LSTM encoding \cite{augenstein-etal-2016-stance}, attention mechanisms \cite{du-ijcai2017-557} or memory networks \cite{mohtarami-etal-2018-automatic}. 
Some neural network models also incorporate lexical features \cite{RiedelASR17,hanselowski-etal-2018-retrospective}. None of these approaches leverage knowledge acquired from massive external corpora. 

\textbf{Approach and Contributions:}
To overcome the limitations of prior works, we present STANCY, a neural network model for stance classification. Given an input pair of a {claim} and a user's {perspective}, our model predicts whether the perspective is \textit{supporting} or \textit{opposing} the claim. For example, the claim \textit{``You have nothing to worry about surveillance, if you have done nothing wrong"} is {supported} by the user perspective \textit{``Information gathered through surveillance could be used to fight terrorism"} and {opposed} by another user perspective \textit{``With surveillance, the user privacy will go away!"}. 

Our model for stance classification leverages representations from the BERT (Bidirectional Encoder Representations from Transformers) neural network model \cite{devlin2018bert}. BERT is trained on huge text corpora and serves as background knowledge. We fine-tune BERT for our task which also allows us to jointly model claims and perspectives.
Furthermore, we enhance our model by augmenting it with a novel consistency constraint to capture agreement between the claim and perspective. 

Key contributions of this paper are:
\squishlist
\item \textbf{Model}: A neural network model for stance classification leveraging BERT representations learned over massive external corpora and a novel consistency constraint to jointly model claims and perspectives. 
\item \textbf{Interpretability}: A simple approach to interpret the contribution of perspective tokens in deciding their stance towards the claim.
\item \textbf{Experiments}: Experiments on a recent dataset, \textit{Perspectrum}, highlighting the effectiveness of our approach with error analysis. 
\squishend

\section{BERT-based Approaches}
In this section, first we describe the base model, BERT\textsubscript{BASE}, that is adapted for the stance classification \cite{CKYCR19}. Thereafter, we present our consistency-aware model, BERT\textsubscript{CONS}.

\subsection{Adapting BERT for Stance Classification}

\label{sec:bert_base}
The goal of the stance classification task is to determine the stance of the user \textit{Perspective ($P$)} with respect to the \textit{Claim ($C$)}. 
Since this task involves a pair of sentences ($C$ and $P$), we follow the approach for sentence pair classification task as proposed in \citet{devlin2018bert, CKYCR19}. 

In order to obtain the representation $X^{P|C}$ of $P$ with respect to $C$, this sentence pair is fused into a single input sequence by using a special classification token ({\tt [CLS]}) and a separator token ({\tt [SEP]}): {\tt[CLS]}~$C_{toks}$~{\tt [SEP]}~$P_{toks}$~{\tt [SEP]}. 
The input sequences are tokenized using WordPiece tokenization. The final hidden state representation corresponding to the {\tt[CLS]} token is used as $X^{P|C} \in R^H$. The classification probability is given by passing this representation through the softmax layer:
\vspace{-0.1cm}
\begin{equation}
\hat{y} = softmax(X^{P|C}W^T) 
\label{eq:final}
\end{equation}
where softmax layer weights $W \in R^{H \times K}$ and $K$ is the number of stance (classification) labels. All the parameters of BERT and $W$ are fine-tuned jointly by minimizing the cross-entropy loss ($loss_{ce}$). The architecture of this model, BERT\textsubscript{BASE}, is shown in Figure~\ref{fig:1a}.

\subsection{Consistency-aware Stance Classification}
In this setting, we want to incorporate the consistency between the claim ($C$) and perspective ($P$) representations. We hypothesize that the latent representations of claim and perspective should be {\em dissimilar} if the perspective \textit{opposes} the claim, whereas their representations should be \textit{similar} if the claim is \textit{supported} by the perspective. We capture this with the following components. 

\textbf{Claim Representation:} To capture the latent representation of the claim, we use only the claim text as the input sequence to BERT, i.e., {\tt[CLS]}~$C$~{\tt [SEP]}. The final hidden state of the first input token ({\tt[CLS]}) is used as the claim's representation $X^{C} \in R^H$.

\textbf{Perspective Representation:} Latent representation of the perspective (with respect to the claim) is captured by fusing the two sequences as described in Section~\ref{sec:bert_base}. We pack the claim and perspective pair as a single input sequence and use the final hidden state of the first input token as the perspective representation $X^{P|C} \in R^H$.

\textbf{Capturing Consistency:} To incorporate the consistency between claim and perspective representations, we use the \textit{cosine embedding loss}:
\vspace{-0.2cm}
\begin{equation*}
\small
loss_{cos} = \begin{cases}
1 - \cos(X^C, X^{P|C}) &  \text{$y_{sim} = 1$} \\
max(0, \cos(X^C, X^{P|C})) & \text{$y_{sim} = -1$} \\
\end{cases}
\end{equation*}
where $\cos(.)$ is the cosine similarity function. $y_{sim}$ is equal to $1$ if the perspective is \textit{supporting} the claim (\textit{similar} representations), and $-1$ if the claim is \textit{opposed} by the perspective (\textit{dissimilar} representations). 

\textbf{Joint Loss:} The classification probabilities are determined by concatenating $X^{P|C}$ and $\cos(X^C, X^{P|C})$ and passing it through a softmax layer. However, unlike the BERT\textsubscript{BASE} configuration, parameters of the consistency-aware model are learned by optimizing the joint loss function: $loss = loss_{ce} + loss_{cos}$. With this joint loss function, we enforce
consistency between latent representations of the claim and perspective. The architecture of this consistency-aware model, BERT\textsubscript{CONS}, is shown in Figure~\ref{fig:1b}.

\section{Experimental Setup}
For our experiments, we consider the base version of BERT\footnote{BERT implementation: \url{https://git.io/fhbJQ}} with 12 layers, 768 hidden size, and 12 attention heads. We fine-tune BERT-based models using the Adam optimizer with learning rates $\{1,3,5\} \times 10^{-5}$ and training batch sizes $\{24, 28, 32\}$. We choose the best parameters based on the development split of the dataset. For measuring the performance, we use per-class and macro-averaged Precision/Recall/F1.

\begin{table}[t]
	\centering
	\small
	\begin{tabular}{C{0.9cm}C{1.9cm}C{1.6cm}C{1cm}}
		\toprule
		\textbf{Split} & \textbf{Supporting Pairs} & \textbf{Opposing Pairs} & \textbf{Total Pairs} \\
		\midrule
		train & 3603 & 3404 & 7007  \\
		dev & 1051 & 1045 &  2096 \\
		test & 1471 & 1302 &  2773 \\
		\midrule
		\textbf{Total} & 6125 & 5751 & 11876 \\
		\bottomrule
	\end{tabular}
	\caption{Perspectrum data statistics.}\label{tab:data_stats}
	\vspace{-0.2cm}
\end{table}

\subsection{Dataset}
We evaluate our approach on the \textit{Perspectrum} dataset \cite{CKYCR19}. \textit{Perspectrum} contains claims and users' perspectives from various online debate websites like {\tt idebate.com}, {\tt debatewise.org}, and {\tt procon.org}. Each claim has different perspectives along with the stance (\textit{supporting} or \textit{opposing} the claim). 
We use the same train/dev/test split as provided in the released dataset. 
Statistics of the dataset is shown in Table~\ref{tab:data_stats}.

\subsection{Baselines}
We use the following baselines:
\squishlistnobullet
	\item \textbf{LSTM}: A long short-term memory (LSTM) model, in which we pass the claim and perspective word representations (using GloVE-6B word embeddings of size 300) through a bidirectional LSTM. Then we concatenate the final hidden states of the claim and perspective, and pass it through dense layers with ReLU activations.
	\item \textbf{ESIM}: An enhanced sequential inference model (ESIM) for natural language inference proposed in \citet{chen-etal-2017-enhanced}.
    \item \textbf{MLP}: Multi-layer perceptron (MLP) based model using lexical and similarity-based features -- presented as a \textit{simple but tough-to-beat baseline} for stance detection in \citet{RiedelASR17}.
    \item \textbf{WordAttn}: Our implementation of word-by-word attention-based model using long short-term memory networks \cite{RocktaschelGHKB15}.
    \item \textbf{LangFeat}: A random forest classifier using linguistic lexicons like NRC lexicon
    \cite{Mohammad:2010:EEC:1860631.1860635}, hedges (e.g., \textit{possibly}, \textit{might}, etc.), positive/negative sentiment words
    \cite{Hu:2004:MSC:1014052.1014073}, MPQA subjective lexicon
    \cite{Wilson:2005:RCP:1220575.1220619} and bias lexicon \cite{recasens-etal-2013-linguistic} along with sentiment scores as features.
    \item \textbf{BERT\textsubscript{BASE}}: Approach proposed in \citet{CKYCR19} (as described in Section~\ref{sec:bert_base}).
    \item \textbf{Human}: Human performance on this task as reported in \citet{CKYCR19}.
\squishend

\begin{table*}[t]
	\centering
	\resizebox{\linewidth}{!}{
		\begin{tabular}{L{2cm}C{1.4cm}C{1.4cm}C{1.4cm}|C{1.4cm}C{1.4cm}C{1.4cm}|C{1.4cm}C{1.4cm}C{1.4cm}}
			\toprule
			\multirow{2}{*}{\textbf{\textbf{Approach}}} & \multicolumn{3}{c|}{\textbf{Supporting}} & \multicolumn{3}{c|}{\textbf{Opposing}} & \multicolumn{3}{c}{\textbf{Overall (Macro)}} \\
			\cmidrule{2-4} \cmidrule{5-7} \cmidrule{8-10}
			& \textbf{Prec.} & \textbf{Recall} & \textbf{F}1 & \textbf{Prec.} & \textbf{Recall} & \textbf{F1} & \textbf{Prec.} & \textbf{Recall} & \textbf{F1} \\
			\midrule \midrule
			LSTM & 63.42 & 58.80 & 61.02 & 56.99 & 61.67 & 59.24 & 60.20 & 60.24 & 60.13 \\
			ESIM & 64.38 & 61.32 & 62.81 & 58.53 & 61.67 & 60.06 & 61.46 & 61.50 & 61.44 \\
			MLP & 64.53 & 60.98 & 62.71 & 58.50 & 62.14 & 60.26 & 61.51 & 61.56 & 61.48 \\ 
			WordAttn & 64.43 & 63.43 & 63.93 & 59.40 & 60.45 & 59.92 & 62.07 & 62.03 & 62.04 \\
			LangFeat & 63.74 & 75.05 & 68.94 & 64.75 & 51.77 & 57.53 & 64.24 & 63.41 & 63.23 \\				
			BERT\textsubscript{BASE} & 78.43 & 80.08 & 79.25 & 76.95 & \textbf{75.12} & 76.02 & 77.69 & 77.60 & 77.63 \\
			\midrule			
			BERT\textsubscript{CONS} & \textbf{79.05} & \textbf{84.64} & \textbf{81.75} & \textbf{81.14} & 74.65 & \textbf{77.76} & \textbf{80.09} & \textbf{79.65} & \textbf{79.95} \\
			\midrule \midrule
			Human & - & - & - & - & - & - & 91.3 & 90.6 & 90.9 \\
			\bottomrule
		\end{tabular}
	}
	\caption{Comparison of our approach BERT\textsubscript{CONS} with different baseline models for stance classification.}
	\label{tab:overall_result}
\end{table*}

\section{Results and Discussion}
Stance classification performance of our model and the baselines on the \textit{test} split of the Perspectrum dataset are presented in Table~\ref{tab:overall_result}. Our consistency-aware model BERT\textsubscript{CONS} outperforms all the other baselines. It achieves a performance improvement of about 2 points in F1-score over the strong baseline corresponding to the BERT\textsubscript{BASE} model (p-value of $4.985\mathrm{e}{-4}$ as per the McNemar test). This highlights the value addition achieved by incorporating consistency cues.
Since the BERT-based models incorporate the knowledge acquired from massive external corpora, our model, BERT\textsubscript{CONS}, captures better semantics and outperforms the other baselines. 

\subsection{Interpreting Token-level Contribution}
Due to the massive structure of BERT with a complex attention mechanism, it is difficult to interpret the significance of different lexical units in the text. Therefore, we propose a simple technique to interpret the contribution of each token in the text in determining the stance.

Given the claim ($C$) and perspective ($P$) pair, we tokenize $P$ into {\em phrases}. We record the change in stance classification probabilities by adding one perspective phrase at a time to the input: 

{\small
$\Delta_{i}=|BERT_{CONS}(C, P_i) - BERT_{CONS}(C, P_{i-1})|$ 
}

\noindent where $P_i$ is the prefix of $P$ up to the $i^{th}$ phrase.
This helps us in understanding the contribution of each perspective phrase towards determining the stance -- the larger the change in the classification probabilities, the larger the contribution. 
For this analysis, we consider unigrams and chunks from a shallow parser as phrases. The top contributing phrases for the \textit{supporting} and \textit{opposing} classes are shown in Table~\ref{tab:top_features}.

\begin{table}[t]
	\centering
	\resizebox{\linewidth}{!}{
		\begin{tabular}{C{4.1cm}C{4.2cm}}
			\toprule
			\textbf{Opposing Class} & \textbf{Supporting Class} \\
			\midrule
			unauthorized, falsely, even though, unlike, cannot, not everyone, could strike, could further weaken, jeopardize, impacts, may not provide, ... &  enabling, ensuring, prevail, positive discrimination, gains, help reduce, would improve, right, would allow, encourage, more effective, ...\\
			\bottomrule
		\end{tabular}
	}
	\caption{Top phrases for determining stance.}
	\label{tab:top_features} 
\end{table}

\subsection{Error Analysis}
In this section, we analyze why the task of stance classification is challenging and why the performance of the best model configuration is far from human performance as observed by the performance gap in Table~\ref{tab:overall_result}.

\noindent
\textbf{Negations:} One of the major challenges in solving this task is understanding negations and their scope. For example, given the claim \textit{``College education is worth it"}, the perspective \textit{``Many college graduates are employed in jobs that \textbf{do not} require college degrees"} is {opposing} the claim. However, our model is not able to capture that the negation phrase \textit{`do not require'} opposes the claim. On the other hand, the presence of negation in the perspective does not necessarily imply that it is {opposing} the claim. Contrast this with the claim \textit{``Chess must be at the Olympics"} and perspective \textit{``Chess is currently \textbf{not} an Olympic sport, but it should be"} -- where the negation is merely a part of the statement and the stance is given by the discourse segment following {\em `but'}.

\noindent
\textbf{Commonsense:} Determining the stance may require commonsense knowledge. For example, the claim \textit{``Chess must be at the Olympics"} is {opposed} by the perspective \textit{``Olympic sports are supposed to be \textbf{physical}"}. To understand this, the model should have the background knowledge that chess is not a physical sport.

\noindent
\textbf{Semantics:} Understanding the stance also involves a deeper understanding of semantics. For example, given the claim \textit{``Make all museums free of charge"} is {opposed} by the perspective \textit{``State funding should be used \textbf{elsewhere}"}. Here, the word \textit{`elsewhere'} is the key cue which determines the stance. However, the presence of the word \textit{`elsewhere'} does not necessarily imply that the perspective is opposing the claim. For instance, the perspective \textit{``We could spend the money \textbf{elsewhere}"} is {supporting} the claim \textit{``The EU should significantly reduce the amount it spends on agricultural production subsidies"}. Hence, the polarity of the word \textit{`elsewhere'} is determined by the context and semantics of the statement.

\section{Conclusion}
In this work, we propose a consistency-aware neural network model for stance classification. Our model leverages representations from the BERT model trained over massive external corpora and a novel consistency constraint to jointly model claims and perspectives. 
Our experiments on a recent benchmark highlight the advantages of our approach. We also study the gap in human performance and the performance of the best model for stance classification.

\section*{Acknowledgments}
This research was partly supported by the ERC Synergy Grant ``imPACT" (No. 610150).

\bibliography{emnlp-ijcnlp-2019}

\begin{thebibliography}{29}
\expandafter\ifx\csname natexlab\endcsname\relax\def\natexlab#1{#1}\fi

\bibitem[{Anand et~al.(2011)Anand, Walker, Abbott, Fox~Tree, Bowmani, and
  Minor}]{anand-etal-2011-cats}
Pranav Anand, Marilyn Walker, Rob Abbott, Jean~E. Fox~Tree, Robeson Bowmani,
  and Michael Minor. 2011.
\newblock \href {https://www.aclweb.org/anthology/W11-1701} {Cats rule and dogs
  drool!: Classifying stance in online debate}.
\newblock In \emph{Proceedings of the 2nd Workshop on Computational Approaches
  to Subjectivity and Sentiment Analysis ({WASSA} 2.011)}, pages 1--9,
  Portland, Oregon. Association for Computational Linguistics.

\bibitem[{Augenstein et~al.(2016)Augenstein, Rockt{\"a}schel, Vlachos, and
  Bontcheva}]{augenstein-etal-2016-stance}
Isabelle Augenstein, Tim Rockt{\"a}schel, Andreas Vlachos, and Kalina
  Bontcheva. 2016.
\newblock \href {https://doi.org/10.18653/v1/D16-1084} {Stance detection with
  bidirectional conditional encoding}.
\newblock In \emph{Proceedings of the 2016 Conference on Empirical Methods in
  Natural Language Processing}, pages 876--885, Austin, Texas. Association for
  Computational Linguistics.

\bibitem[{Baly et~al.(2018)Baly, Mohtarami, Glass, M{\`a}rquez, Moschitti, and
  Nakov}]{baly-etal-2018-integrating}
Ramy Baly, Mitra Mohtarami, James Glass, Llu{\'\i}s M{\`a}rquez, Alessandro
  Moschitti, and Preslav Nakov. 2018.
\newblock \href {https://doi.org/10.18653/v1/N18-2004} {Integrating stance
  detection and fact checking in a unified corpus}.
\newblock In \emph{Proceedings of the 2018 Conference of the North {A}merican
  Chapter of the Association for Computational Linguistics: Human Language
  Technologies, Volume 2 (Short Papers)}, pages 21--27, New Orleans, Louisiana.
  Association for Computational Linguistics.

\bibitem[{Bar-Haim et~al.(2017)Bar-Haim, Bhattacharya, Dinuzzo, Saha, and
  Slonim}]{bar-haim-etal-2017-stance}
Roy Bar-Haim, Indrajit Bhattacharya, Francesco Dinuzzo, Amrita Saha, and Noam
  Slonim. 2017.
\newblock \href {https://www.aclweb.org/anthology/E17-1024} {Stance
  classification of context-dependent claims}.
\newblock In \emph{Proceedings of the 15th Conference of the {E}uropean Chapter
  of the Association for Computational Linguistics: Volume 1, Long Papers},
  pages 251--261, Valencia, Spain. Association for Computational Linguistics.

\bibitem[{Chen et~al.(2017)Chen, Zhu, Ling, Wei, Jiang, and
  Inkpen}]{chen-etal-2017-enhanced}
Qian Chen, Xiaodan Zhu, Zhen-Hua Ling, Si~Wei, Hui Jiang, and Diana Inkpen.
  2017.
\newblock \href {https://doi.org/10.18653/v1/P17-1152} {Enhanced {LSTM} for
  natural language inference}.
\newblock In \emph{Proceedings of the 55th Annual Meeting of the Association
  for Computational Linguistics (Volume 1: Long Papers)}, pages 1657--1668,
  Vancouver, Canada. Association for Computational Linguistics.

\bibitem[{Chen et~al.(2019)Chen, Khashabi, Yin, Callison-Burch, and
  Roth}]{CKYCR19}
Sihao Chen, Daniel Khashabi, Wenpeng Yin, Chris Callison-Burch, and Dan Roth.
  2019.
\newblock \href {https://www.aclweb.org/anthology/N19-1053} {Seeing things from
  a different angle:discovering diverse perspectives about claims}.
\newblock In \emph{Proceedings of the 2019 Conference of the North {A}merican
  Chapter of the Association for Computational Linguistics: Human Language
  Technologies, Volume 1 (Long and Short Papers)}, pages 542--557. Association
  for Computational Linguistics.

\bibitem[{Chen and Ku(2016)}]{chen-ku-2016-utcnn}
Wei-Fan Chen and Lun-Wei Ku. 2016.
\newblock \href {https://www.aclweb.org/anthology/C16-1154} {{UTCNN}: a deep
  learning model of stance classification on social media text}.
\newblock In \emph{Proceedings of {COLING} 2016, the 26th International
  Conference on Computational Linguistics: Technical Papers}, pages 1635--1645,
  Osaka, Japan. The COLING 2016 Organizing Committee.

\bibitem[{Devlin et~al.(2019)Devlin, Chang, Lee, and
  Toutanova}]{devlin2018bert}
Jacob Devlin, Ming-Wei Chang, Kenton Lee, and Kristina Toutanova. 2019.
\newblock Bert: Pre-training of deep bidirectional transformers for language
  understanding.
\newblock In \emph{Proceedings of the 2019 Conference of the North American
  Chapter of the Association for Computational Linguistics: Human Language
  Technologies}.

\bibitem[{Du et~al.(2017)Du, Xu, He, and Gui}]{du-ijcai2017-557}
Jiachen Du, Ruifeng Xu, Yulan He, and Lin Gui. 2017.
\newblock \href {https://doi.org/10.24963/ijcai.2017/557} {Stance
  classification with target-specific neural attention}.
\newblock In \emph{Proceedings of the Twenty-Sixth International Joint
  Conference on Artificial Intelligence, {IJCAI-17}}, pages 3988--3994.

\bibitem[{Ferreira and Vlachos(2016)}]{ferreira-vlachos-2016-emergent}
William Ferreira and Andreas Vlachos. 2016.
\newblock \href {https://doi.org/10.18653/v1/N16-1138} {{E}mergent: a novel
  data-set for stance classification}.
\newblock In \emph{Proceedings of the 2016 Conference of the North {A}merican
  Chapter of the Association for Computational Linguistics: Human Language
  Technologies}, pages 1163--1168, San Diego, California. Association for
  Computational Linguistics.

\bibitem[{FNC-1(2016)}]{FNC}
FNC-1. 2016.
\newblock \href {http://www.fakenewschallenge.org/} {Fake news challenge stage
  1 (fnc-1): Stance detection}.

\bibitem[{Hanselowski et~al.(2018)Hanselowski, PVS, Schiller, Caspelherr,
  Chaudhuri, Meyer, and Gurevych}]{hanselowski-etal-2018-retrospective}
Andreas Hanselowski, Avinesh PVS, Benjamin Schiller, Felix Caspelherr, Debanjan
  Chaudhuri, Christian~M. Meyer, and Iryna Gurevych. 2018.
\newblock \href {https://www.aclweb.org/anthology/C18-1158} {A retrospective
  analysis of the fake news challenge stance-detection task}.
\newblock In \emph{Proceedings of the 27th International Conference on
  Computational Linguistics}, pages 1859--1874, Santa Fe, New Mexico, USA.
  Association for Computational Linguistics.

\bibitem[{Hasan and Ng(2013)}]{hasan-ng-2013-stance}
Kazi~Saidul Hasan and Vincent Ng. 2013.
\newblock \href {https://www.aclweb.org/anthology/I13-1191} {Stance
  classification of ideological debates: Data, models, features, and
  constraints}.
\newblock In \emph{Proceedings of the Sixth International Joint Conference on
  Natural Language Processing}, pages 1348--1356, Nagoya, Japan. Asian
  Federation of Natural Language Processing.

\bibitem[{Hasan and Ng(2014)}]{Hasan2014WhyAY}
Kazi~Saidul Hasan and Vincent Ng. 2014.
\newblock Why are you taking this stance? identifying and classifying reasons
  in ideological debates.
\newblock In \emph{EMNLP}.

\bibitem[{Hu and Liu(2004)}]{Hu:2004:MSC:1014052.1014073}
Minqing Hu and Bing Liu. 2004.
\newblock \href {https://doi.org/10.1145/1014052.1014073} {Mining and
  summarizing customer reviews}.
\newblock In \emph{Proceedings of the Tenth ACM SIGKDD International Conference
  on Knowledge Discovery and Data Mining}, KDD '04, pages 168--177, New York,
  NY, USA. ACM.

\bibitem[{Kochkina et~al.(2017)Kochkina, Liakata, and
  Augenstein}]{KochkinaLA17}
Elena Kochkina, Maria Liakata, and Isabelle Augenstein. 2017.
\newblock \href {http://arxiv.org/abs/1704.07221} {Turing at semeval-2017 task
  8: Sequential approach to rumour stance classification with branch-lstm}.
\newblock \emph{CoRR}, abs/1704.07221.

\bibitem[{Lukasik et~al.(2016)Lukasik, Srijith, Vu, Bontcheva, Zubiaga, and
  Cohn}]{lukasik-etal-2016-hawkes}
Michal Lukasik, P.~K. Srijith, Duy Vu, Kalina Bontcheva, Arkaitz Zubiaga, and
  Trevor Cohn. 2016.
\newblock \href {https://doi.org/10.18653/v1/P16-2064} {{H}awkes processes for
  continuous time sequence classification: an application to rumour stance
  classification in twitter}.
\newblock In \emph{Proceedings of the 54th Annual Meeting of the Association
  for Computational Linguistics (Volume 2: Short Papers)}, pages 393--398,
  Berlin, Germany. Association for Computational Linguistics.

\bibitem[{Mohammad et~al.(2016)Mohammad, Kiritchenko, Sobhani, Zhu, and
  Cherry}]{mohammad-etal-2016-semeval}
Saif Mohammad, Svetlana Kiritchenko, Parinaz Sobhani, Xiaodan Zhu, and Colin
  Cherry. 2016.
\newblock \href {https://doi.org/10.18653/v1/S16-1003} {{S}em{E}val-2016 task
  6: Detecting stance in tweets}.
\newblock In \emph{Proceedings of the 10th International Workshop on Semantic
  Evaluation ({S}em{E}val-2016)}, pages 31--41, San Diego, California.
  Association for Computational Linguistics.

\bibitem[{Mohammad and Turney(2010)}]{Mohammad:2010:EEC:1860631.1860635}
Saif~M. Mohammad and Peter~D. Turney. 2010.
\newblock \href {http://dl.acm.org/citation.cfm?id=1860631.1860635} {Emotions
  evoked by common words and phrases: Using mechanical turk to create an
  emotion lexicon}.
\newblock In \emph{Proceedings of the NAACL HLT 2010 Workshop on Computational
  Approaches to Analysis and Generation of Emotion in Text}, CAAGET '10.

\bibitem[{Mohtarami et~al.(2018)Mohtarami, Baly, Glass, Nakov, M{\`a}rquez, and
  Moschitti}]{mohtarami-etal-2018-automatic}
Mitra Mohtarami, Ramy Baly, James Glass, Preslav Nakov, Llu{\'\i}s M{\`a}rquez,
  and Alessandro Moschitti. 2018.
\newblock \href {https://doi.org/10.18653/v1/N18-1070} {Automatic stance
  detection using end-to-end memory networks}.
\newblock In \emph{Proceedings of the 2018 Conference of the North {A}merican
  Chapter of the Association for Computational Linguistics: Human Language
  Technologies, Volume 1 (Long Papers)}, pages 767--776, New Orleans,
  Louisiana. Association for Computational Linguistics.

\bibitem[{Recasens et~al.(2013)Recasens, Danescu-Niculescu-Mizil, and
  Jurafsky}]{recasens-etal-2013-linguistic}
Marta Recasens, Cristian Danescu-Niculescu-Mizil, and Dan Jurafsky. 2013.
\newblock \href {https://www.aclweb.org/anthology/P13-1162} {Linguistic models
  for analyzing and detecting biased language}.
\newblock In \emph{Proceedings of the 51st Annual Meeting of the Association
  for Computational Linguistics (Volume 1: Long Papers)}, pages 1650--1659,
  Sofia, Bulgaria. Association for Computational Linguistics.

\bibitem[{Riedel et~al.(2017)Riedel, Augenstein, Spithourakis, and
  Riedel}]{RiedelASR17}
Benjamin Riedel, Isabelle Augenstein, Georgios~P. Spithourakis, and Sebastian
  Riedel. 2017.
\newblock \href {http://arxiv.org/abs/1707.03264} {A simple but tough-to-beat
  baseline for the fake news challenge stance detection task}.
\newblock \emph{CoRR}, abs/1707.03264.

\bibitem[{Rockt{\"{a}}schel et~al.(2016)Rockt{\"{a}}schel, Grefenstette,
  Hermann, Kocisk{\'{y}}, and Blunsom}]{RocktaschelGHKB15}
Tim Rockt{\"{a}}schel, Edward Grefenstette, Karl~Moritz Hermann, Tom{\'{a}}s
  Kocisk{\'{y}}, and Phil Blunsom. 2016.
\newblock \href {http://arxiv.org/abs/1509.06664} {Reasoning about entailment
  with neural attention}.
\newblock In \emph{ICLR}.

\bibitem[{Sobhani et~al.(2017)Sobhani, Inkpen, and
  Zhu}]{sobhani-etal-2017-dataset}
Parinaz Sobhani, Diana Inkpen, and Xiaodan Zhu. 2017.
\newblock \href {https://www.aclweb.org/anthology/E17-2088} {A dataset for
  multi-target stance detection}.
\newblock In \emph{Proceedings of the 15th Conference of the {E}uropean Chapter
  of the Association for Computational Linguistics: Volume 2, Short Papers},
  pages 551--557, Valencia, Spain. Association for Computational Linguistics.

\bibitem[{Somasundaran and Wiebe(2010)}]{somasundaran-wiebe-2010-recognizing}
Swapna Somasundaran and Janyce Wiebe. 2010.
\newblock \href {https://www.aclweb.org/anthology/W10-0214} {Recognizing
  stances in ideological on-line debates}.
\newblock In \emph{Proceedings of the {NAACL} {HLT} 2010 Workshop on
  Computational Approaches to Analysis and Generation of Emotion in Text},
  pages 116--124, Los Angeles, CA. Association for Computational Linguistics.

\bibitem[{Sridhar et~al.(2015)Sridhar, Foulds, Huang, Getoor, and
  Walker}]{sridhar-etal-2015-joint}
Dhanya Sridhar, James Foulds, Bert Huang, Lise Getoor, and Marilyn Walker.
  2015.
\newblock \href {https://doi.org/10.3115/v1/P15-1012} {Joint models of
  disagreement and stance in online debate}.
\newblock In \emph{Proceedings of the 53rd Annual Meeting of the Association
  for Computational Linguistics and the 7th International Joint Conference on
  Natural Language Processing (Volume 1: Long Papers)}, pages 116--125,
  Beijing, China. Association for Computational Linguistics.

\bibitem[{Sun et~al.(2018)Sun, Wang, Zhu, and Zhou}]{sun-etal-2018-stance}
Qingying Sun, Zhongqing Wang, Qiaoming Zhu, and Guodong Zhou. 2018.
\newblock \href {https://www.aclweb.org/anthology/C18-1203} {Stance detection
  with hierarchical attention network}.
\newblock In \emph{Proceedings of the 27th International Conference on
  Computational Linguistics}, pages 2399--2409, Santa Fe, New Mexico, USA.
  Association for Computational Linguistics.

\bibitem[{Walker et~al.(2012)Walker, Anand, Abbott, and
  Grant}]{walker-etal-2012-stance}
Marilyn Walker, Pranav Anand, Rob Abbott, and Ricky Grant. 2012.
\newblock \href {https://www.aclweb.org/anthology/N12-1072} {Stance
  classification using dialogic properties of persuasion}.
\newblock In \emph{Proceedings of the 2012 Conference of the North {A}merican
  Chapter of the Association for Computational Linguistics: Human Language
  Technologies}, pages 592--596, Montr{\'e}al, Canada. Association for
  Computational Linguistics.

\bibitem[{Wilson et~al.(2005)Wilson, Wiebe, and
  Hoffmann}]{Wilson:2005:RCP:1220575.1220619}
Theresa Wilson, Janyce Wiebe, and Paul Hoffmann. 2005.
\newblock \href {https://doi.org/10.3115/1220575.1220619} {Recognizing
  contextual polarity in phrase-level sentiment analysis}.
\newblock In \emph{Proceedings of the Conference on Human Language Technology
  and Empirical Methods in Natural Language Processing}, HLT '05, pages
  347--354, Stroudsburg, PA, USA. Association for Computational Linguistics.

\end{thebibliography}
\bibliographystyle{acl_natbib}

\end{document}